\documentclass{article} %

\usepackage{authblk}

\usepackage{amsmath,amsfonts,bm}

\def\eqref#1{equation~\ref{#1}}

\def\1{\bm{1}}

\DeclareMathAlphabet{\mathsfit}{\encodingdefault}{\sfdefault}{m}{sl}
\SetMathAlphabet{\mathsfit}{bold}{\encodingdefault}{\sfdefault}{bx}{n}

\usepackage[utf8]{inputenc} %
\usepackage[T1]{fontenc}    %

\usepackage[margin=1in]{geometry}
\usepackage{natbib}

\usepackage[hidelinks,breaklinks]{hyperref}       
\hypersetup{colorlinks=true, citecolor=DarkBlue, linkcolor=darkblue, urlcolor=darkblue}

\usepackage{url}            %
\usepackage{booktabs}       %
\usepackage{amsfonts}       %
\usepackage{nicefrac}       %
\usepackage{microtype}      %
\usepackage{xcolor}  %
\definecolor{DarkGreen}{rgb}{0.1,0.5,0.1}
\definecolor{DarkRed}{rgb}{0.5,0.1,0.1}
\definecolor{DarkBlue}{rgb}{0.1,0.1,0.5}
\definecolor{darkblue}{rgb}{0, 0, 0.5}

\usepackage{amsmath}
\usepackage{amsthm}
\usepackage{amssymb}
\usepackage{cleveref}
\usepackage{graphicx}

\usepackage{placeins}

\usepackage{algorithm}

\usepackage{stmaryrd}
\usepackage{subfig}

\usepackage{times}
\usepackage{latexsym}

\usepackage[T1]{fontenc}

\usepackage[utf8]{inputenc}

\usepackage{microtype}

\usepackage{inconsolata}

\usepackage{graphicx}

\usepackage{booktabs}
\usepackage{amsmath}

\usepackage{amssymb}
\usepackage{tikz}

\usepackage[normalem]{ulem}
\usepackage{xcolor}

\usepackage{tcolorbox}
\usepackage{amsthm}

\usepackage{ifthen}
\usepackage{enumitem}

\usepackage{algpseudocode}

\newcommand{\igraf}[3]{
\ifthenelse
 {\equal{#1}{h}}
  {\includegraphics[height=#2\textheight]{#3}}
  {\includegraphics[width=#2\textwidth]{#3}}
}

\newcommand{\xhdr}[1]{\paragraph{\bf #1.}}

\newcommand{\omt}[1]{}

\newcommand\supproof[1]{}

\def\univ{{\cal X}}
\def\coll{{\cal L}}

\def\trueL{{K}}
\def\harmL{{H}}
\def\colltrue{{\cal L}_{\trueL}}
\def\collharm{{\cal L}_{\harmL}}

\newtheorem{theorem}{Theorem}[section]
\newtheorem{corollary}{Corollary}[section]
\newtheorem{definition}{Definition}

\title{Safe Language Generation in the Limit}

\date{}

\author{Antonios Anastasopoulos, Giuseppe Ateniese, Evgenios M. Kornaropoulos\\
Department of Computer Science \\
George Mason University\\
  \texttt{\{antonis,ateniese,evgenios\}@gmu.edu} \\}

\begin{document}

\maketitle

\begin{abstract}
Recent results in learning a language in the limit have shown that, although language identification is impossible, language generation is tractable. 
As this foundational area expands, we need to consider the implications of language generation in real-world settings. 

This work offers the first theoretical treatment of \textit{safe} language generation. Building on the computational paradigm of learning in the limit, we formalize the tasks of safe language identification and generation. We prove that under this model, safe language identification is impossible, and that safe language generation is at least as hard as (vanilla) language identification, which is also impossible. Last, we discuss several intractable and tractable cases.
\end{abstract}

\section{Introduction}
\label{sec:intro}

As large language models (LLMs) are increasingly deployed in high-stakes settings, safety failures can lead to significant social, economic, and security harms. 
These alarming failures have been reported in news articles, lawsuits, and academic works. 

Multiple media reports and lawsuits allege that prolonged, emotionally intimate interactions with LLMs reinforced suicidal ideation~\citep{6abc_raine_suit_2025} and delusional thinking~\citep{wsj_suicide_delusions_2025} rather than de-escalating crises, with claims that the system validated harmful thoughts~\citep{people_shamblin_2025}, discouraged seeking human support, or failed to redirect users to appropriate crisis resources. These allegations include wrongful-death cases involving adolescents and young adults~\citep{pbs_chatgpt_dangerous_2025} and a separate case involving the amplification of conspiratorial beliefs linked to a murder-suicide~\citep{wsj_murder_suicide_2025}. Thus, unsafe LLM behavior is not merely hypothetical but does arise in practice.

Academic and industrial research efforts have largely, if not entirely, focused on \emph{heuristic} approaches on benchmarks and mitigation. 
There is no dearth of LLM safety benchmarks focusing on various safety aspects, such as truthfulness~\citep{lin-etal-2022-truthfulqa}, toxicity~\citep{hartvigsen-etal-2022-toxigen}, unethical scenarios~\citep{shen2024anything}, or generally adversarial dialogues~\citep{ganguli2022red}. Another research direction focuses on mitigation strategies ranging from pre-training data filtering to unlearning of undesirable behaviors, and from direct prompting to post-training for safety.\footnote{We direct the interested reader to the plethora of relevant surveys that synthesize recent works on benchmarks and mitigation strategies~\cite[][\textit{inter alia}]{liu2025safety,li2025security,yong2025state,
anthropic2025agentic}.}

Beyond empirical research, however, what is desperately needed is research to understand the foundational computational task of LLM safety, particularly as LLMs gain greater agency and are applied across diverse languages and contexts.

\xhdr{Our Contributions} Our work offers the first foundational treatment of \textit{safe} language generation: 

\begin{enumerate}[noitemsep,nolistsep,leftmargin=*]
    \item We introduce a generic and versatile abstraction that captures the task of \emph{safe language generation in the limit} (\S\ref{sec:model}) under the  learning paradigm of~\citet{GOLD1967447}. 
    In our new model, the learner sees examples from the desired language as well as examples from the harmful language, and the goal is to produce unseen examples from the (safe) set difference between the two languages. 
    \item As a first step, we introduce the task of \emph{safe language identification in the limit}, where the goal of the learner is not to produce an unseen example but rather pinpoint the safe language among a collection of languages.
    We prove that safe language identification is impossible~(\S\ref{sec:idisimpossible}). 

    \item We then prove that safe language generation \emph{is at least as hard} as language identification in the limit~(\S\ref{sec:sg}), and discuss several \emph{intractable and tractable cases} of safe language generation~(\S\ref{sec:tractable}).
\end{enumerate}

\noindent The implications of our work are rather surprising.
Seminal results from computational learning theory have shown that language generation in the limit is, in a formal sense, fundamentally easier than broader learning tasks
such as language identification. 
In fact, (vanilla) language generation \emph{is possible} even against an adversary that presents positive training examples in a worst-case manner. 
In contrast, our work shows for the first time that \textit{safe} generation is a fundamentally harder task. 
We note that our work does not aim to invalidate empirical research on LLM safety; rather, it showcases the fundamental complexity of the task, which may inform practical decisions on safety trade-offs.

\section{Related Work}

\xhdr{Language \emph{Identification} in the Limit}
Gold's foundational paper \citep{GOLD1967447} introduced language identification in the limit: there is a countable domain $\univ$ and a countable collection of candidate languages $\coll=\{L_1,L_2,\ldots\}$, one of which is the unknown target language $K$. 
From an enumeration of positive examples of the true language $K$, the learner $A$ must output in every round an exact description of $K$ (often an index in $\coll$). 
Crucially, the learner does not get feedback on the correctness of their output. 
Gold showed that identification from positive data alone is impossible in general. \citet{ANGLUIN1980117} gave a characterization of when identification from positive data \emph{is} possible, via finite ``telltale'' sets that allow the learner to rule out competing hypotheses, and \citet{angluin1988identifying} studied identification when examples are drawn stochastically. 
Together, these results expose a core obstacle: any finite prefix may be consistent with multiple distinct languages in the limit.

\xhdr{Language \emph{Generation} in the Limit} 
\citet{DBLP:conf/nips/KleinbergM24} introduced language generation in the limit, replacing identification by a weaker goal: after some finite time, the learner must generate previously \emph{unseen} elements of $K$. Their main (constructive) theorem shows that for every countable collection $\coll$ over a countable domain, there exists an algorithm that generates in the limit from every $K\in\coll$. 
In particular, this establishes a \emph{separation} between identification and generation, i.e., generation can be universally achievable even when identification is not. Details of the \texttt{KM} algorithm are available in Appendix~\ref{app:km}.

\xhdr{On Refined Notions of Success} 
After establishing that unseen-example generation is tractable, the natural question becomes which additional requirements preserve this tractability and which do not. 
\citet{pmlr-v291-raman25a} take a more learning theory-driven approach and provide a characterization of hypothesis classes that are uniformly and
non-uniformly generatable. 
\citet{DBLP:conf/stoc/KalavasisMV25,kalavasis2024,pmlr-v291-charikar25a} characterize the tension between consistency and expressiveness, which is typically seen in practice as the trade-off between hallucination and mode collapse.  
\citet{kleinbergwei2025density,kleinbergwei2025partialenumeration} propose density measures as a quantitative proxy for breadth and study how such guarantees behave under partial enumeration. 
\citet{hanneke2025on} showed that it is possible to have collections $L_1$ and $L_2$ that are generatable individually, while their union $L_1 \cup L_2$ is not.  \citet{arenas2025complexity} emphasizes that, despite theoretical guarantees that language generation is possible from positive examples alone, simple formal language classes may require samples beyond any computable bound to generate correctly.
Another line of work keeps the basic generation goal but varies the \emph{information model}.
\citet{pmlr-v267-raman25a} study generation when the observed examples contain a bounded amount of adversarial corruption. \citet{bai2026noiselossfeedback} unify noise, loss, and feedback within a single framework.
\citet{mehrotra2025contamination} extends robustness to regimes of unbounded contamination.

\xhdr{Why Is This Work Different}  
Across these variants, there is still a single target language $\trueL$, and an output is ``wrong'' only because it falls outside $\trueL$. 
Our work introduces an additional constraint, motivated by safety concerns in LLM deployment; that is, we model allowable outputs as the set difference $\trueL \setminus \harmL$, where $\harmL$ is a \emph{harmful language} revealed over time along with the true language $\trueL$. The key distinction is that the learner needs to generate while avoiding strings in $\harmL$, which 
can overlap with $\trueL$, i.e., a string can be consistent with the target language while still being disallowed.

\section{Safe Language Identification is Impossible}
\label{sec:idisimpossible}

Before considering safely \emph{generating} in the limit, we start with a more fundamental question: \emph{Can we \emph{identify} the safe language (i.e., $\trueL\setminus\harmL$) in the limit?}
Given the negative result from~\cite{GOLD1967447}, which shows that language identification from positive examples alone is impossible, the question above appears hopeless. 
On a second thought, one can see the strings from $\univ$ that are labeled as members of $\harmL$ as \emph{negative examples} with respect to the safe language $\trueL\setminus\harmL$.
Specifically, strings that will only be labeled with $0$ are strings that are in $\harmL$ but not in $\trueL$ (thus, they are not in $\trueL\setminus\harmL$ either), while strings that are labeled as both $0$ and $1$ are in $\trueL\cap\harmL$ (thus, they are not in $\trueL\setminus\harmL$).

It is plausible that this new perspective may lead to tractability, since the inclusion of negative examples is known to make otherwise intractable identification learning problems solvable with both positive and negative examples~\cite{GOLD1967447}. 
We note here that the sufficiency and impossibility theorems of \citet{ANGLUIN1980117} on language identification in the limit \emph{do not carry over verbatim} in our setting.
This is because when a string from $\trueL$ is revealed by the adversary, it does not immediately serve as a positive example, since it may later be revealed to belong in $\harmL$.

\begin{definition}[\textbf{Safe Identification $\mathcal{SI}$}]
Fix some $\trueL$ from the language collection  $\colltrue$ and some $\harmL$ from the language collection  $\collharm$. 
Add language $\trueL\setminus\harmL$ to a random location of $\colltrue$. 
In each step $t$, the algorithm $A$ observes a labeled set $S_t$ such that each string is labeled $1$ if it is a member of $\trueL$ and $0$ if it is a member of $\harmL$. 
At each timestep, $A$ outputs an index $i$ of $\colltrue$. 
We say that algorithm A \textbf{safely identifies from $\trueL$ given a harmful $\harmL$ in the limit} if there is some time $t'\in \mathbb{N}$ such that, 
for all $t> t'$ the algorithm's guess at step $t$ is $i$ for which $L_i=\trueL\setminus\harmL$ with respect to $\colltrue$.
\end{definition}

In the above definition, the language $\trueL\setminus\harmL$ is inserted at a random location within $\colltrue$ to ensure the algorithm $A$ has no prior knowledge of its position in the collection $\colltrue$.

\begin{theorem} 
For any safe language identification algorithm $A^\mathcal{SI}$, there exists a countable $\univ$, a pair of language collections $\colltrue$ and $\collharm$, and a choice of $\trueL$ and $\harmL$, such that $A^\mathcal{SI}$ can not safely identify from $\trueL$ given a harmful $\harmL$ in the limit. 
\label{thm:identificationImpossibility}
\end{theorem}

\begin{proof}
In this proof, we construct the collections of languages $\colltrue$ and $\collharm$ such that no matter how $A$ attempts to identify the safe language, there is always a choice of $\trueL$ and $\harmL$ that they can not identify in the limit. 

Our construction operates over the universe $\univ=\mathbb{Z}$ of integers, and the languages will be defined as arithmetic progressions. 
One can ``relabel'' each integer with a unique string; thus, the impossibility result is not tied only to languages of integers.

\begin{figure*}[t]
\includegraphics[width=12.5cm]{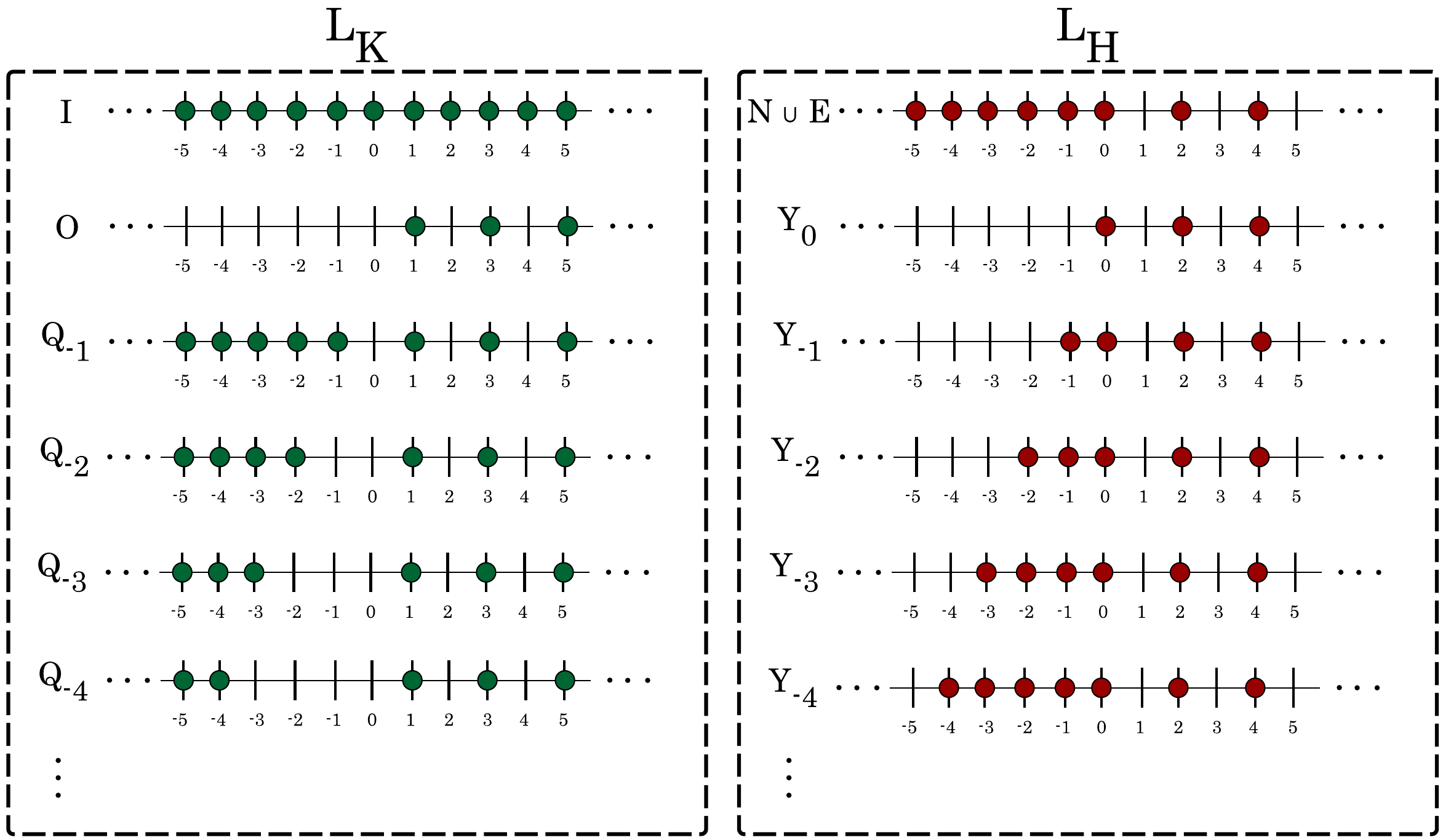}
\centering
\caption{An illustration of the collection of $\colltrue$ and $\collharm$ for the constructed instantiation for which there is no timestep in which an algorithm identifies the correct language $\trueL \setminus \harmL$ consistently.
Let $I$ denote the set of integers, $E$ the set of even integers, and $O$ the set of odd integers.} 
\label{fig:safeID}
\end{figure*}

\xhdr{The Languages} To set up the language collections $\colltrue$ and $\collharm$, we define several languages that we will need in the proof. 
Let $O$ be the language of odd positive integers and $E$ be the language of even positive integers:
\begin{align*}
O=&\{2i-1:i=1,2,\ldots\},\\
E=&\{2i:i=0,1,2,\ldots\}.
\end{align*}

Next, we define two more languages that are parametrized so as to generate a countable collection of languages. 
The first language is $Y_{-a}$, which, on a high level, contains the negative integers from $-a$ all the way to $0$ (i.e., $-a\rightarrow 0$) as well as all the even positive integers. Formally, it is defined as
\begin{displaymath}
Y_{-a}=\{-a+i:i=0,\ldots,a\}\cup E\text{, for $a\geq 0$} 
\end{displaymath}
The second language is $Q_{-b}$ that contains negative integers from $-b$ to $-\infty$ (i.e., $-\infty\rightarrow -b$) as well as all the positive odd integers. Formally:
\begin{displaymath}
Q_{-b}=\{-b-i:i=0,1,2,\ldots\}\cup O\text{, for $b\geq 1$}.
\end{displaymath}
Finally, we define all integers as
$I=\{i:i=\ldots,-2,-1,0,1,2,\ldots\}$ and the negative integers as $N=\{-i:i=1,2,\ldots\}$.

\xhdr{The Language Collections} We can now define the language collections for the safe identification instance. An illustration of these collections is provided in Figure~\ref{fig:safeID}.  

\begin{itemize}
\item $\collharm$: The collection $\collharm$ contains language $N\cup E$ as well as all parameterizations of languages $Y_{-a}$ for $a\geq 0$. Thus, it is a countable collection. 
\item $\colltrue$: The collection $\colltrue$ is the union of language $I$, language $O$, and all the parameterizations of languages $Q_{-b}$ for $b\geq 1$. Thus, it is a countable collection.
\end{itemize}

Now that we have established the language collections, we will show that for any identification algorithm $A$ there exists ($i$) a choice of $\trueL\in\colltrue$, ($ii$) a choice of $\harmL\in\collharm$, and ($iii$) a labeled enumeration $E$ of  $\trueL$ and $\harmL$, such that if  $\trueL\setminus\harmL$ is the safe language and the adversary provides the enumeration $E$ to algorithm $A$, then the algorithm $A$ can not safely identify in the limit. 

We will construct the enumeration in phases and adaptively select $\trueL$ and $\harmL$ based on the outputs of $A$. 
The proposed construction of $E$ proceeds in multiple and possibly infinite phases. 

We define the following two labeled sequences that we will use throughout the proof: $T_{1}=\big((0,1)$, $(1,1)$, $(-1,1)$, $(2,1)$, $(-2,1)$, $(3,1)$, $(-3,1)$, $\ldots\big)$, where the second element of each pair is the label $1$ that signifies that the number is in $\trueL$; and $T_{2}=\big((0,0), (2,0), (4,0),\ldots\big)$, where the second element of the pair is the label $0$ that signifies that the number is in $\harmL$. 
Notice that the elements in $T_{2}$ belong to \emph{all} potential harmful languages in $\collharm$ and there are infinite such elements. 
Finally, we define a sequence $T$ that interleaves the labeled examples from $T_1$ and $T_2$, i.e., $T=\left(T_1(1),T_2(1), T_1(2),T_2(2),T_1(3),T_2(3),\ldots\right)$. 
Looking ahead, the adversary will use the sequence $T$ to reveal elements to the algorithm $A$, observe its identification guess, and adapt the final enumeration $E$ accordingly.

\xhdr{Phase 1 of Construction of Enumeration $E$} 
In the first phase, we construct the enumeration $E$ by revealing elements as if $\trueL = I$ and $\harmL = Y_0$. 
Specifically, at any round $t$, the adversary presents the labeled element $T(t)$ to the algorithm $A$. 
The behavior of algorithm $A$ is unknown; thus, it either correctly identifies $\trueL \setminus \harmL$ as $Q_{-1}$ or not. 

Next, we consider the following case analysis about $A$'s behavior: ($i$) either there is some finite step $t_1\in\mathbb{N}$ such that the guess of the algorithm $A$ is $Q_{-1}$, or ($ii$) there is no $t_1\in\mathbb{N}$ such that $A$ guesses $Q_{-1}$. 
In the latter case, the adversary commits to the choices $\trueL = I$ and $\harmL = Y_0$ and enumeration $E=T$ and is guaranteed that at no point in time the algorithm $A$ identifies $\trueL\setminus\harmL$ correctly, which concludes the proof. 
In the former case, the adversary adapts the enumeration by revealing in step $t_1+1$ the harmful labeled example $(-1, 0)$ to the algorithm. 
Given this new labeled pair, the previous inference of $A$ in which $\harmL$ was hypothesized to be $Y_0$ is clearly wrong since $-1\notin Y_0$. 
Next, the construction transitions to Phase 2. 

\xhdr{Phase 2 of Construction of $E$} 
Specifically, at each round $t > t_1 + 1$, the adversary enumerates the labeled element $T(t - 1)$ to the algorithm $A$, i.e., continuing the enumeration of elements from the sequence $T$ starting from where phase $1$ ended.

Next, we consider two cases: ($i$) either there is some finite step $t_2>t_1 + 1$ such that the guess of the algorithm $A$ for $\trueL \setminus \harmL$ is $Q_{-2}$, or ($ii$) there is no $t_2\in\mathbb{N}$ such that $A$ guesses $Q_{-2}$. 
In the latter case, the adversary commits to the choices $\trueL = I$ and $\harmL = Y_{-1}$ and enumeration $E=\left(T(1:t_1),(-1,0),T(t_1+1:\infty)\right)$ 
and it is guaranteed that at no point in time the algorithm $A$ guesses $\trueL\setminus\harmL$ correctly, which concludes the proof. 
In the former case, the adversary adapts the enumeration by revealing in step $t_2+1$, the harmful labeled example $(-2, 0)$ to the algorithm. 
Given this new labeled pair, the previous inference of $A$ in which $\harmL$ was $Y_{-1}$ is clearly wrong since $-2\notin Y_{-1}$. 
The construction then progresses to Phase 3.

\xhdr{Inductive Argument} Generalizing the strategy of the first two phases, the adversary continues building the target enumeration inductively. In case there is any phase $l\in \mathbb{N}$ in which the case ($ii$) of the case analysis is true, then the impossibility result holds for this algorithm $A$, and the proof is over. 
In case there is no such phase $l\in \mathbb{N}$, then the adversary constructed an enumeration $E$ of labeled examples for the languages $\trueL=I$ and $\harmL=N\cup E$. 
We highlight here that  $\trueL\setminus\harmL=I\setminus (N\cup E)=O$ is in $\colltrue$, therefore both choices of the adversary come from the appropriate language collection, and the safe language is in $\colltrue$. 
In this case, the algorithm $A$ guesses incorrectly for infinitely many steps $t\in\mathbb{N}$. 
Finally, the enumeration $E$ for this case is built inductively over all phases, i.e., 
\begin{align*}
    E=& \big( T(1:t_1), (-1,0), T(t_1+1:t_2), (-2,0),\\ &T(t_2+1:t_3), (-3,0), \ldots \big) 
\end{align*}
\end{proof}

\xhdr{Discussion} This result implies that even when safe language identification has access to harmful words from the target language, their presence alone is insufficient to guarantee consistent and correct identification. 
Or to put it differently, a subset of negative examples (words that belong to both $\trueL$ and $\harmL$) is just as uninformative for identification as having no negative examples at all. 

\noindent The main reason examples from $\harmL$ do not make safe language identification easier (despite being negative examples) is that the set of strings observed by $A^\mathcal{SI}$ does not necessarily include all possible negative examples. 
That is, it is possible to have strings that are neither in $\trueL\setminus\harmL$ nor in $\harmL$. 
Overall, \citet{GOLD1967447} showed that inference from positive data is strictly less powerful than inference from positive and negative data; the above theorem shows that with just a subset of negative examples (as opposed to exhaustive enumeration), the inferential power of the learner does not increase.

\section{The Safe Generation Problem}
\label{sec:model}

Let $\colltrue=\{L_1,L_2,L_3,\ldots\}$ be a countable collection of candidate \emph{true} languages, each an infinite subset of a countable universe $\univ$ of strings.
Let $\collharm=\{L_1,L_2,L_3,\ldots\}$ be a countable collection of infinite candidate \emph{harmful} languages, also subsets of $\univ$. 
Both $\colltrue$ and $\collharm$ are allowed repetitions, i.e., multisets. 
An adversary and an algorithm (or else the learner) $A$ play the following game:
The adversary secretly selects a true language $\trueL\in\colltrue$ and a harmful language $\harmL\in\collharm$. At each timestep $t$, it presents a string from either $\trueL$ or $\harmL$, i.e., a labeled string $(w_t,y_t)$, where label is $y_t=1$ if $w_t\in\trueL$ and $y_t=0$ if $w_t\in\harmL$. 
In each round, the adversary reveals a string from either the true or the harmful language. 
We call $\trueL\setminus\harmL$ the \emph{safe} language and $\perp$ the bottom string. 
Additionally, through this generation, the adversary commits to an \emph{enumeration} of $\trueL$  and an \emph{enumeration} of $\harmL$. 
The adversary can repeat strings
in these enumerations, but for every string $w \in \trueL$ and every string $w \in \harmL$, there must be at least one time step $t$ in which it appears.
Let $S_t$ denote the set of strings that the adversary has enumerated in steps
$1$ through $t$. 
At the end of each round, $A$ guesses unseen words from the safe language but receives no feedback on whether the guesses are correct.

\begin{definition}[\textbf{Safe Generation $\mathcal{SG}$}]
Fix some $\trueL$ from the language collection  $\colltrue$ and some $\harmL$ from the language collection  $\collharm$. 
In each step $t=1,2,\ldots$ the algorithm $A$ observes a labeled set $S_t$ such that each string is labeled $1$ if it is a member of $\trueL$ and $0$ if it is a member of $\harmL$. 
At each timestep, $A$ outputs a string $a_t$ (its guess for an unseen string in $\trueL\setminus\harmL$). 
We say that algorithm A \textbf{safely generates from $\trueL$ given a harmful $\harmL$ in the limit} if 
for all labeled enumerations of $\trueL$ and $\harmL$, 
$(i)$ in case $\trueL\setminus\harmL$ is countably infinite, there is some time $t'\in \mathbb{N}$ such that for all steps $t> t'$ the algorithm's guess $a_t$  belongs to $(\trueL\setminus\harmL)\setminus S_t$, and $(ii)$ in case $\trueL\setminus\harmL$ is either finite or empty, there is some time $t''\in \mathbb{N}$, for which for all steps $t> t''$ the algorithm's guess is $\perp$, where $\perp\notin\univ$. 
\label{def:generation}
\end{definition}

\xhdr{On the Emptiness of Set Difference} In the above definition for safe generation, we expect two distinct behaviors of the algorithm depending on whether the set difference $\trueL\setminus\harmL$ is infinite or not. 
In case it is infinite,  the algorithm is expected to generate (in the limit) strings from the infinite set $\trueL\setminus\harmL$. 
In case it is finite or empty, it is impossible\footnote{In this case, the impossibility is due to a lack of enough safe strings to generate in the limit.} to generate from $\trueL\setminus\harmL$ in the limit, and the algorithm needs to identify this case by outputting the special symbol $\perp$. 
This asymmetry is intentional: since safety is a strict requirement in our setting, emitting unsafe strings when no safe output exists would be unacceptable. The symbol $\perp$ explicitly signals the absence of any safe generative behavior.

\begin{table}[t]
    \centering
    \footnotesize
    \begin{tabular}{r p{3.2cm} | r p{3.2cm} | r p{3.2cm}}
    \toprule
    \multicolumn{2}{c|}{\small{\textbf{Computational Problems}}} &
    \multicolumn{2}{c|}{\small{\textbf{Setup}}} &
    \multicolumn{2}{c}{\small{\textbf{Algorithm Related}}} \\
    \midrule
    $\mathcal{LI}$ & Language Identification &
    $\univ$ & Countable universe of strings &
    $A^\mathcal{P}$ & Algorithm solving $\mathcal{P}$ \\
    $\mathcal{LG}$ & Language Generation &
    $\coll, \colltrue, \collharm$ & Collections of languages &
    $t$ & Timestep $t$ \\
    $\mathcal{SI}$ & Safe Language Identification &
    $K$ & True language of $\mathcal{LI}$ or $\mathcal{LG}$ &
    $w_t$ & Word revealed at timestep $t$ \\
    $\mathcal{SG}$ & Safe Language Generation &
    $\trueL, \harmL$ & True, harmful languages  &
    $S_t$ & Set of words revealed up to $t$ \\
     &  &
     & for $\mathcal{SI}$ or~$\mathcal{SG}$ &
    $\mathcal{C}_t$ & Languages consistent w. $S_t$ \\
    & & $\trueL\setminus\harmL$ &Safe language of $\mathcal{SI}$ or~$\mathcal{SG}$ &
    $\perp$ & Bottom string \\
    \bottomrule
    \end{tabular}
    \caption{Notation summary.}
    \label{tab:notation}
\end{table}

\xhdr{Why a Direct Application of \texttt{KM} Fails}
A naive attempt to apply the \texttt{KM} algorithm for $\mathcal{LG}$ (see Appendix~\ref{app:km}) directly to $\mathcal{SG}$ fails. 
To see why, consider substituting generation from the difference with generation solely from $\colltrue$, while discarding candidate languages whenever they become inconsistent due to $\harmL$'s examples. 
That is, whenever a previously shown word $w\in\trueL$ is also enumerated in $\harmL$, a modified \texttt{KM} algorithm could recompute 
the set of consistent languages and continue as usual. 
However, this naive approach breaks a crucial core invariant underlying the correctness arguments of \texttt{KM}. 
In the \texttt{KM} setting for $\mathcal{LG}$, when the correct language ($\trueL$) is first considered, it will \textit{remain consistent} for all timesteps thereafter. 
However, under the $\mathcal{SG}$ setting, any word $w\in\trueL\cap\harmL$ that has only been revealed (so far) as a member of $\trueL$ and not $\harmL$, renders the correct language ($\trueL\setminus\harmL$) inconsistent with $S_t$, because $\forall w \in \trueL\cap\harmL: w \not\in \trueL\setminus\harmL$. 
Thus, the correct language $\trueL\setminus\harmL$ is not guaranteed to remain consistent across all possible enumerations.

\section{Safe Language Generation is \textit{at Least} as Hard as Language Identification}
\label{sec:sg}

In this section, we will show that safe generation in the limit from a language $\trueL$ given a harmful $\harmL$ is at least as hard as language identification in the limit. 
Our argument assumes an algorithm $A^\mathcal{SG}$ that solves any safe generation problem and uses it to construct a new algorithm $A^\mathcal{LI}$ that solves any instance of the language identification problem. 
Thus, if there were an easy way to solve $\mathcal{SG}$, we would have an easy way to solve $\mathcal{LI}$.

\begin{theorem}\label{thm:idtosg}
    Let $A^{\mathcal{SG}}$ be an algorithm that safely generates from a language $\trueL$ given a harmful language $\harmL$ in the limit for any enumeration of the two languages from collections $\colltrue$ and $\collharm$ from domain $\univ$. 
    Then, for any collection $\mathcal{L}$ from the domain $\univ$, there exists an algorithm $A^{\mathcal{ID}}$ that uses $A^{\mathcal{SG}}$ as a subroutine to identify $K$ in the limit given any enumeration of any $K \in \mathcal{L}$.
\end{theorem}

\xhdr{High-Level Proof Sketch} 
At a high level, Algorithm $A^{\mathcal{ID}}$ at time-step $t$ will consider the first $t$ languages from $\mathcal{L}$, and will maintain the  set $\mathcal{C}_t$ of languages that are consistent with the set $S_t$ of revealed strings. After some timestep $z$, the true $L_z=K$ will be inserted into the set of consistent languages $\mathcal{C}_t$ and never removed from it. 

Consider all languages $L_{i\leq z}$ up to language $L_z$.
In the limit, any such language that is a strict subset of $L_z$ will eventually be ruled out, since the enumeration will produce a string in $L_z$ that is absent from it. 
The only languages $L_i$ remaining in the list (with $i<z$), in the limit, will be any superset languages $L_i$, i.e., $L_i \supset K $.

Given such a list of consistent languages, $A^{\mathcal{ID}}$ will take advantage of the asymmetry in the outputs of the safe generation algorithm $A^{\mathcal{SG}}$. 
In particular, for all languages $L_i$ consistent with $S_t$, $A^{\mathcal{ID}}$ will construct an instance of a safe generation problem with a new $L^*_i$ enumerated as the safe language (using a new finite enumeration), and with a new $K^*$ enumerated as the harmful language (using $S_t$).

Specifically, $A^{\mathcal{ID}}$ defines the new languages $L^*_i = \{\langle (w,n) \rangle : w \in L_i, n \in \mathbb{N}^+\} $, and analogously for $K^*$; where $\langle \cdot \rangle$ denotes a fixed computable string encoding of the enclosed pair as a string. 
Given any enumeration $G_t$ of up to $t$ elements from $L_i$, $A^{\mathcal{ID}}$ uses it to construct an enumeration $G^*_t = \{\langle(w,n)\rangle : w \in G_t, 1\leq n \leq t\}$ for the new language $L^*_i$. 
Analogously, $A^{\mathcal{ID}}$ constructs an enumeration $S^*_t = \{\langle (w,n) \rangle : w \in S_t, 1\leq n \leq t\}$ of $K^*$. 
Note that both $G^*_t$ and $S^*_t$ are valid enumerations of the respective languages, since $G_t$ and $S_t$ are valid enumerations.

Notice that for any superset language $L_i$, i.e., $K \subseteq L_i$, there are two cases for the outputs of safe generation from $L_i\setminus K$. 
Either $L_i\setminus K$ is infinite, or $L_i\setminus K$  is finite (recall the case $L_i\setminus K=\emptyset$ is handled in the limit through consistency).
If $L_i\setminus K$ is infinite, then $L^*_i\setminus K^*$ is also infinite, and thus the $A^\mathcal{SG}$ algorithm will find some word to return in the limit.
If $L_i\setminus K$ is finite, then $L^*_i\setminus K^*$ is again infinite (as it contains the following infinite elements: $L^*_i\setminus K^* = \{\langle(w,n)\rangle : w \in L_i \setminus K, n \in \mathbb{N}^+\}$). 
We emphasize that, counterintuitively, even though $L_i \setminus K$ is finite and is supposed to be exhausted in the limit, there are still infinitely many safe strings $\langle (w, n) \rangle$ in $L^*_i \setminus K^*$, 
as each $w \in L_i \setminus K$ gives rise to infinitely many encodings by pairing with distinct natural numbers. 
Thus, in this case as well, the $A^\mathcal{SG}$ algorithm in the limit will find some word to return. 

But for the language $L_z$, safe generation from $L^*_z \setminus K^*$ will be impossible (since $L_z=K$ and thus $L^*_z=K^*$ as well).
The guarantee of hypothesized $A^{\mathcal{SG}}$ for safe generation is that there exists some $t \in \mathbb{N}$ after which the returned word $w_t \in \trueL \setminus \harmL$, \textit{if possible}. 
In the limit, the safe generation algorithm will correctly return $\perp$ only for language $L_z$, and it will also find some word $w_i\not=\perp$ for all other superset languages $L_i: K \subseteq L_i, i < z$.
Thus, if we return the index $i$ of the smallest language that is (a) consistent, i.e.,  $L_i \in \mathcal{C}_t$ and (b) for which $A^\mathcal{SG}$ returned $w_i = \perp$, then in the limit, we will always return $z$, solving the language identification problem.  \hfill$\square$

\xhdr{Solving Safe Generation $\Longrightarrow$ Language Identification}

We will now elaborate, showing the complete proof that safe generation in the limit from a language $\trueL$ given a harmful $\harmL$ is at least as hard as language identification in Angluin's setting.
As a reminder, our proof assumes an algorithm $A^\mathcal{SG}$ that solves the safe generation and uses it to construct an algorithm $A^\mathcal{LI}$ that solves any instance of the language identification problem. 
Thus, if there were an easy way to solve $\mathcal{SG}$, we would have an easy way to solve $\mathcal{LI}$, which is impossible.

\begin{algorithm}[t]
\small
\caption{$A^\mathcal{LI}$: Language Identification from Safe Generation}\label{alg:li-from-sg}
\begin{algorithmic}[1]
\algrenewcommand\algorithmicrequire{\textbf{Inputs:}}
\Require
\Statex \quad $\bullet$ Domain $\mathcal{X}$
\Statex \quad $\bullet$ Language collection $\coll$
\Statex \quad $\bullet$ Enumeration $E=\bigl(w_1, w_2,\ldots\bigr)$ of $K$
\Statex \quad $\bullet$ Membership oracle for $\mathcal{O}(w,i)$, where $L_i\in \coll$. \Comment{\texttt{Returns \texttt{True} if $w \in L_i$, else \texttt{False}.}}
\Statex \quad $\bullet$ Safe Generation algorithm $A^\mathcal{SG}(E_l,\colltrue,\collharm)$, where $E_l$ is a labeled enumeration of $\trueL,\harmL$ and $\colltrue,\collharm$ are language collections.

\For{$t = 1, 2, \dots$}
    \State Let $S_t = \left\{w_1,\ldots,w_t\right\}$ be the sequence of observed words from (unknown) $\trueL$.
    \State Compute the \textit{consistent} set
    \begin{align*}
        \mathcal{C}_t = \{L_i \in \coll : S_t \subseteq L_i \text{ and } i \leq t \}.
    \end{align*}
    \State Enumerate domain prefix $\mathcal{X}_t=\{x_1,\ldots,x_t\}$.
    \For{i=1,\ldots,t}
        \State Obtain an enumeration of $L_i$:
        \begin{align*}
            G_{L_i} = \{x\in\mathcal{X}_t : x\in L_i\}.
        \end{align*}
        \State Obtain an enumerations of the vertical padding $L^*_i$:
        \begin{align*}
            G^*_{L_i} = \{\langle(x,n)\rangle : x\in G_{L_i}, 1 \leq n \leq t \}.
        \end{align*}
        \State Obtain an enumerations of the vertical padding $S^*_t$:
        \begin{align*}
            S^*_{t} = \{\langle(x,n)\rangle : x\in S_{t}, 1 \leq n \leq t \}.
        \end{align*}
        \State Create labeled enumeration $E^*_i$ concatenating $G^*_{L_i}$ (as true) and $S^*_t$ (as harmful):
        \begin{align*}
            E^*_i = \{(x,1) : x\in G^*_{L_i} \} \ \cup \{ (x,0): x\in S^*_t\}
        \end{align*}
        \State Feed $E^*_i,\coll^*$ to $A^{\mathcal{SG}}$ to obtain output $w^t_i = A^\mathcal{SG}(E^*_i,\coll^*,\coll^*)$.
    \EndFor
    \State Let
    \begin{align*}
        \mathcal{N} = \{i \leq t : L_i \in \mathcal{C}_t \text{ and } w^t_i=\perp\}. 
    \end{align*}
    \If{$\mathcal{N}\not=\emptyset$}
        \State Let $z' = \min \{i \in \mathcal{N}\}$.
        \State \Return $z'$ \Comment{Output the index of the identified language}
    \Else
        \State \Return 1 \Comment{Returns arbitrary index.}
    \EndIf
\EndFor
\end{algorithmic}
\end{algorithm}

\begin{proof}
Let $z\in\mathbb{N}$ be the smallest number such that $L_z=K$. 
Our algorithm outputs the language with the \textit{smallest} index that satisfies the two conditions described above. 
Thus, to get the desired result we need to show that all languages in $\coll_{z-1}=\{L_1,L_2,\ldots,L_{z-1}\}$ that precede $L_z$ do \textit{not} satisfy these conditions (in the limit, for all sufficiently large $t$), while the target language $L_z$ does satisfy these conditions (again, in the limit).
To that end, we divide $\coll_z$ into two disjoint subsets: $L_{z-1}^{\supset} = \{L\in\coll_{z-1}: L \supset L_z\}$ and  $L_{z-1}^{\not\supset} = \{L \in \coll_{z-1}: L \not\supset L_z\}$.
In other words, $L_{z-1}^{\supset}$ is the set of all languages that precede $L_z$ and are strict supersets of it,
and $L_{z-1}^{\not\supset}$ is the set of all languages that precede $L_z$ and are \textit{not} strict supersets of it.
Since $L_z \not\in \coll_{z-1}$, we have that $L_{z-1}^{\not\supset} \cup L_{z-1}^{\supset} = \coll_{z-1}$. 
We now handle the two sets separately.

We first consider the set $L_{z-1}^{\not\supset}$.
By definition, for every such language $L_{i_j}$ in this subset, there exists some element $x_{i_j} \in L_z, x_{i_j}\not\in L_{i_j}$.
Moreover, since the adversary presents a complete enumeration of $L_z$, there exists some timestep $t_{l_j}$ such that $w_{t_{l_j}} = x_{i_j}$ where this word is enumerated.
We define $t^*_1 = \max_{j\leq k}t_{l_j}$.
Using the definition of the consistent set $\mathcal{C}_t$, one can see that for all $t\geq t^*_1$ these languages are not consistent with $S_t$, i.e., $L_{z-1}^{\not\supset} \not\in \mathcal{C}_t$.

We now turn our focus to set $L_{z-1}^{\supset}$. Let $L_{j_1},\ldots,L_{j_m}$ be the languages of this set, where $0\leq m\leq z-1$.
Recall that these are the languages that are strict supersets of the correct language $L_z$.
For any $i \leq m$ consider the execution of the safe generation algorithm with input the collection $\coll^*$ of vertically padded languages from $\coll$, an enumeration of $L^*_i$ as the enumeration of the safe language, and the enumeration $S^*_t$ (of the vertical padding of the correct $L_z$) as the enumeration of the harmful language.

Since $L_i \supset L_z$, this implies that $L_i\setminus L_z \not= \emptyset$. 
There are two cases for this set non-empty difference: it is either finite or infinite. 
If $L_i\setminus K$ is infinite, then $L^*_i\setminus K^*$ is also infinite, and thus the $A^\mathcal{SG}$ algorithm will find some word to return in the limit.
If $L_i\setminus K$ is finite, then $L^*_i\setminus K^*$ is again infinite (as it contains the following infinite elements: $L^*_i\setminus K^* = \{(w,n) : w \in L_i \setminus K, n \in \mathbb{N}\}$).
In both cases, thus, there will exist some $t'_{l_i}$ such that for all $t>t'_{l_i}$ the safe generation algorithm will find a word $w\in L^*_i\setminus K^*$ to return, i.e., the safe generation algorithm will (correctly) \textit{not} return $\perp$.
We define $t^*_2 = \max_{i\leq m} t'_{l_i}$.
It follows that for all $t \geq t^*_2$, the safe generation algorithm will return a non-$\perp$ word for all vertically padded languages of all $L_{j_1},\ldots,L_{j_m}$.

Last, let us consider the language $L_z$. Obviously, for $t\geq z$ we have $L_z\in\mathcal{C}_t$.
In other words, after timestep $t = z$, the correct true $K$ will be inserted into the set of consistent languages and never removed from it. 
Unlike all other languages, though, when we call the safe generation algorithm $A^\mathcal{SG}$ with an enumeration of $L^*_z$ as the true language and with $S^*_t$ as the enumeration of the harmful language, there will exist some timestep $t'$ after which $A^\mathcal{SG}$ will always (correctly) return $\perp$. 
This is because $S_t$ is a valid enumeration of $L_z$, and $S^*_t$ is also a valid enumeration of $L^*_z$, and hence the set difference $L^*_z \setminus K^*$ is empty. 
Let $t^*_3$ be $t^*_3 = \max(z,t')$.

We now have all the ingredients we need to prove our result. 
We let $t^* = \max \{t^*_1,t^*_2,t^*_3\}$, and consider any $t \geq t^*$,
By definition of $t^*$, for any language $L_i, i < z$, either $L_i \not\in \mathcal{C}_t$, or the safe generation algorithm $A^\mathcal{SG}$ returns some word $w \not= \perp$, so the two conditions are mutually exclusive and cannot be simultaneously satisfied for this language.
Both conditions \textit{will} be satisfied simultaneously for $L_z$ since $L_z \in \mathcal{C}_t$ and the safe generation algorithm will return $\perp$.
Hence, the smallest indexed language that satisfies both of our conditions is indeed $L_z$.
Thus, our algorithm $A^\mathcal{LI}$ achieves identification in the limit.
\end{proof}

\paragraph{Why Do We Need Vertical Padding?}
We briefly explain why we need to construct the vertical padding languages $L^*_i$ on which to run the safe generation algorithm.
What happens if we run the safe algorithm directly on enumerations of the original $L_i$ and the enumeration $S_t$ of the true language? 
For the cases of superset languages $L_i \supset K$, we will again have two cases for their set difference: if the set difference $L_i \setminus K$ is infinite, then a correct safe generation algorithm will, in the limit, find some word to return. But if the set difference $L_i \setminus K$ is finite, then in the limit the correct safe generation behavior is to output $\perp$.
This latter behavior overlaps with the behavior when the set difference is empty.
Thus, in the limit, we would not be able to use the outputs of the $A^\mathcal{SG}$ algorithm to distinguish between empty and non-empty set differences, which is needed to distinguish between strict superset languages and the true language.

On the other hand, the vertical padding languages have the property than if the set difference of two languages $L_1 \setminus L_2$ is non-empty (either finite or infinite), then the set difference of their vertical paddings $L^*_1 \setminus L^*_2$ will always be infinite (since $\mathbb{N}^+$ is infinite).
If the set difference of two languages $L_1 \setminus L_2$ is empty, then also the set difference of their vertical paddings $L^*_1 \setminus L^*_2$ will always be empty as well.

Thus, by moving to the space of vertically padded languages, we eliminate ambiguity in the behavior of the safe generation algorithm, as we remove the possibility of a finite set difference between two languages, leaving only the possibilities of either empty or infinite.

\FloatBarrier

\section{Tractable and Intractable Cases for Safe Language Generation}
\label{sec:tractable}

In this section, we analyze three variants of our core safe generation problem and determine whether any of them render safe generation tractable. 
First, in Section~\ref{subsec:idisnotenough}, we show that even if both languages $\trueL$ and $\harmL$ are identifiable (with respect to~\citet{ANGLUIN1980117}), the safe generation problem $\mathcal{SG}$ remains impossible in the general case. This impossibility follows from the fact that deciding whether the set difference of two countable sets is empty is undecidable. 
Next, in Section~\ref{subsec:setdifference}, we drop the identifiability assumption on $\trueL$ and $\harmL$, but instead assume $A$ has access to an oracle that can answer whether the set difference of any two sets is empty or not. We show that access to this oracle alone is still insufficient to make $\mathcal{SG}$ tractable. 
Trivially, if $A$ has access to such an oracle \emph{and} both languages are identifiable, then the safe language generation problem is tractable. 
Interestingly, the above two conditions are sufficient but not necessary. 
In the final variant from Section~\ref{subsec:tractable}, we describe a tractable setting: namely, when the set difference between every pair of languages in $\colltrue$ and $\collharm$ is guaranteed to be infinite by construction, while the chosen languages are not necessarily identifiable.

\subsection{Identifiability Alone is Not Enough for Safe Generation in the Limit}
\label{subsec:idisnotenough}

\xhdr{Angluin's Condition for Identifiability}
A natural first direction to explore is what happens when $\colltrue$ and $\collharm$ are designed in such a way that \emph{any choice} of $\trueL$ and $\harmL$ is identifiable in the limit. 
The work of \citet{ANGLUIN1980117} established a necessary and sufficient condition for identifiability. 
 
\vspace{0.05cm}
\begin{tikzpicture}
  \draw node[draw=black,fill=black!8,rounded corners,inner sep=0.5ex,text width=0.95\textwidth] {\small
\textbf{Condition.}~\cite{ANGLUIN1980117} Fix a language collection $\mathcal{L}$. The collection $\mathcal{L}$ is said to satisfy Angluin's condition if for any index $i$, there is a finite set of strings $T_i$ such that $T_i$ is a subset of $L_i\in\mathcal{L}$, i.e., $T_i\subseteq L_i$, and the following holds:
\begin{center}
 For all $j\geq 1$, if $T_i\subseteq L_j$, then $L_j$ is not a proper subset of $L_i$. 
\end{center}
A telltale oracle is a primitive that, given index $i$, outputs an enumeration of the set $T_i$. 
 };
\end{tikzpicture}

 We begin with a decision problem that will anchor the subsequent analysis and show that it is undecidable to safely generate even if $A$ deals with identifiable $\trueL$ and  $\harmL$. 
Language (in the computability sense) $L_{\text{\texttt{DIFF-EMPTY}}}$ contains all the pairs of infinite sets such that their difference is empty. 
%
%
\begin{align*}
L_{\text{\texttt{DIFF-EMPTY}}}
&= \{(X_1,X_2) \mid X_1,X_2 \subseteq \univ,
\quad |X_1|=|X_2|=\infty,\; X_1 \setminus X_2 = \emptyset \}.
\end{align*}

\begin{theorem}
$L_{\text{\texttt{DIFF-EMPTY}}}$ is undecidable.
\label{thm:diffEmpty}
\end{theorem}

\begin{proof}
In this elementary proof, we will show that if $L_{\text{\texttt{DIFF-EMPTY}}}$ is decidable, then we can build a decider for $L_{\text{\texttt{HALT}}}$, which is the language for the halting problem. 
Suppose for the sake of contradiction that $L_{\text{\texttt{DIFF-EMPTY}}}$ is decidable. 
Then, there must be a Turing machine $D$ that is the decider of $L_{\text{\texttt{DIFF-EMPTY}}}$. 
Next, we build a decider $M^*$ for $L_{\text{\texttt{HALT}}}$ by using $D$. 
The TM $M^*$ takes as input the description of a TM $<T>$ and an input string $s$. 
Let $X_1=\mathbb{N}$ be a set whose members are the natural numbers. 
Let $X_2=\{n\in\mathbb{N}: T(s)\text{ does not halt exactly at step }n\}$.
Membership in $X_2$ is decidable: on input $n$, simulate $T(s)$ for exactly $n$ steps and check whether its first halting step is $n$.
If $T(s)$ never halts, then $X_1\setminus X_2=\emptyset$; if $T(s)$ first halts at step $h$, then $X_1\setminus X_2=\{h\}$.
Thus, the TM $M^*$ feeds the above two sets to the decider of $L_{\text{\texttt{DIFF-EMPTY}}}$, namely $D$, and flips its answer. This decides $L_{\text{\texttt{HALT}}}$, a contradiction.
\end{proof}

Suppose, for simplicity, that both $\trueL$ and $\harmL$ are given explicitly to an algorithm $A$ tasked with safe generation.
Even in this setting, $A$ \emph{cannot guarantee that it will output, within a finite number of steps, a string $w$} such that $w \in \trueL$ and $w \notin \harmL$.
The reason is that $A$ may be dealing with an instance $(\trueL,\harmL) \in L_{\text{\texttt{DIFF-EMPTY}}}$, in which case no string satisfying these properties exists.
Because it is undecidable to determine whether a given instance has an empty difference, $A$ cannot determine when to stop seeking a safe generation because none exists; even when explicit descriptions of $\trueL$ and $\harmL$ are provided rather than learned.

\begin{corollary}
There are identifiable collections $\colltrue$ and $\collharm$ such that no algorithm can safely generate in the limit unless \texttt{DIFF-EMPTY} is decidable.
\end{corollary}

\subsection{Access to a Set-Difference Oracle Alone is Not Enough for Safe Generation}
\label{subsec:setdifference}

\begin{figure}
  \centering
  \includegraphics[width=.5\textwidth, trim={0  0 0 0.7cm}, clip]{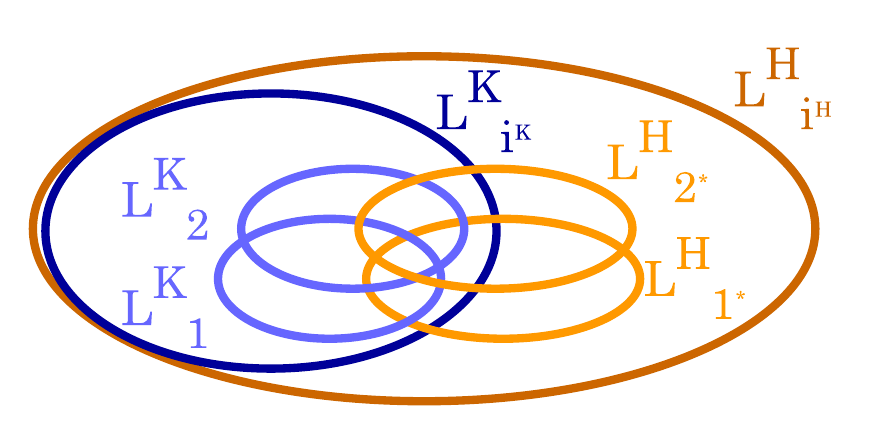}
  \caption{An illustration of collections in which each phase seemingly has infinite safe strings, but the enumerated languages $(L^{\trueL}_{i^{\trueL}},L^{\harmL}_{i^{\harmL}})$ have no safe strings.}
  \label{fig:unsafe}
\end{figure}

In what follows, we assume that the algorithm $A^{\mathcal{SG}}$ is equipped with a set-difference oracle $\mathcal{O}^{SD}(\cdot,\cdot)$, which returns $1$ if the difference between two input infinite collections is empty and $0$ otherwise.
We show that even with access to this oracle, safe generation in the limit is still impossible.
Our argument constructs two collections of languages and an adversarial enumeration that diagonalizes against any fixed algorithm.
Although the algorithm may appear to generate safely at each finite stage, the adversary repeatedly forces hypothesis revisions.
In the limit, the revealed enumerations correspond to a pair of languages with empty difference, for which the only safe output is $\perp$, yet all algorithms can be misled and fail to recognize this setting.
Figure~\ref{fig:unsafe} illustrates this behavior, where the adversarial enumeration switches between pairs $(L^K_1,L^H_{1^*}),(L^K_2,L^H_{2^*}),\ldots$, and in case $A$ pivots its hypothesis, the enumeration can be made from $(L^K_{i^K},L^H_{i^H})$, which has no safe outputs.

\begin{theorem}
There are collections $\colltrue$ and $\collharm$ such that no algorithm can safely generate in the limit even with access to set difference oracle $\mathcal{O}^{SD}$, i.e., \texttt{DIFF-EMPTY}($\cdot$) being decidable.
\label{thm:setDiffOracleImpossible}
\end{theorem}

\begin{proof}
In this proof, we make a diagonalization argument for two language collections that we construct. 
Suppose that the following property holds for $\colltrue=\{L^{\trueL}_1,L^{\trueL}_2,\ldots\}$ and $\collharm=\{L^{\harmL}_1,L^{\harmL}_2,\ldots\}$:

\vspace{0.3cm}
\begin{tikzpicture}
  \draw node[draw=black,fill=black!20,rounded corners,inner sep=0.6ex,text width=0.95\textwidth] {\small
(\textbf{Property $P^*$}) There exists an index $i^{\trueL}$ and $i^{\harmL}$ for which $L^{\trueL}_{i^{\trueL}}\subseteq L^{\harmL}_{i^{\harmL}}$, such that
 for every pair of finite sets of strings $(T^{\trueL},T^{\harmL})$ such that $T^{\trueL}\subset L^{\trueL}_{i^{\trueL}}$ and $T^{\harmL}\subset L^{\harmL}_{i^{\harmL}}$ the following holds:
\begin{center}
 There exists a $j\geq 1$ and an associated $j^*\geq 1$, such that (1) if $T^{\trueL}\subseteq L^{\trueL}_j$, then $L^{\trueL}_j$ is a proper subset of $L^{\trueL}_{i^{\trueL}}$, and (2) if $T^{\harmL}\subseteq L^{\harmL}_{j^*}$, then $L^{\harmL}_{j^*}$ is a proper subset of $L^{\harmL}_{i^{\harmL}}$, and (3) $L^{\trueL}_j\setminus  L^{\harmL}_{j^*}$ has infinite cardinality. 
\end{center}
 };
\end{tikzpicture}

Let $E^{\trueL}_{i^{\trueL}}$ be any enumeration of the $L_{i^{\trueL}}$ and $E^{\harmL}_{i^{\harmL}}$ be any enumeration of $L_{i^{\harmL}}$.  
We will show that for any generator $G$ there exists an adversary that enumerates from  $E^{\trueL}_{i^{\trueL}}$  and $E^{\harmL}_{i^{\harmL}}$ such that $G$ cannot safely generate from $L_{i^{\trueL}}$ in the limit.

\textbf{Phase 1.} Let $x^{\trueL}_1$ be the first element of the enumeration $E^{\trueL}_{i^{\trueL}}$, i.e., $x_1=E^{\trueL}_{i^{\trueL}}(1)$. Set $T^{\trueL}$ to be $\{x^{\trueL}_1\}$.   
Let $x^{\harmL}_1$ be the first element of the enumeration $E^{\harmL}_{i^{\harmL}}$, i.e., $x_1=E^{\harmL}_{i^{\harmL}}(1)$. Set $T^{\harmL}$ to be $\{x^{\harmL}_1\}$.  

From property $P^*$ we know that there must be an $i_1$ and an $i^*_1$ such that
(1) $T^{\trueL}\subseteq L^{\trueL}_{i_1}$ where $L^{\trueL}_{i_1}$ is a proper subset of $L^{\trueL}_{i^{\trueL}}$, and (2) $T^{\harmL}\subseteq L^{\harmL}_{i_1^*}$ where $L^{\harmL}_{i_1^*}$ is a proper subset of $L^{\harmL}_{i^{\harmL}}$, and (3) $L^{\trueL}_{i_1}\setminus  L^{\harmL}_{i^*_1}$ has infinite cardinality.

\textbf{Subphase 1.A:}
Consider an enumeration of $L^{\trueL}_{i_1}$ constructed by traversing the fixed enumeration $E^{\trueL}_{i^{\trueL}}$ and using only the elements of $L^{\trueL}_{i_1}$ that appear in it and transfer them to $E^{\trueL}_{i_1}$ in the same order. 
Consider an enumeration of $L^{\harmL}_{i^*_1}$ constructed by traversing the fixed enumeration $E^{\harmL}_{i^{\harmL}}$ and using only the elements of $L^{\harmL}_{i^*_1}$ that appear in it and transfer them to $E^{\harmL}_{i^*_1}$ in the same order.

We note that the infinite sequence $E^{\trueL}_{i_1}$ (resp. $E^{\harmL}_{i^*_1}$) is indeed an enumeration of $L^{\trueL}_{i_1}$ (resp. $L^{\harmL}_{i^*_1}$) since $L^{\trueL}_{i_1}$ (resp. $L^{\harmL}_{i^*_1}$) is a proper subset of $L_{i^{\trueL}}$ (resp. $L_{i^{\harmL}}$), therefore, all of its strings appear in $E^{\trueL}_{i^{\trueL}}$ (resp. $E^{\harmL}_{i^{\harmL}}$). 
At any round $t$ the adversary presents the string $E^{\trueL}_{i_1}(t)$ and $E^{\harmL}_{i^*_1}(t)$ to the generator $G$.

Consider two cases: ($i$) either there is some finite $t_1\in\mathbb{N}$ such that the generator $G$ will generate an element from $L^{\trueL}_{i_1}\setminus L^{\harmL}_{i^*_1}$, or ($ii$) there is no such $t_1\in\mathbb{N}$. 
Notice that the adversary can verify whether the algorithm is in case ($i$), by observing a generated string $x$ such that the membership oracle of $L^{\trueL}_{i_1}$ says ``member'' while the membership oracle of $\harmL_{j_1}$ says ``non-member''. 
Analogously, the adversary can verify whether the algorithm is in case ($ii$), by observing no generated string $x$ such that the membership oracle of $L^{\trueL}_{i_1}$ says ``member'' and the membership oracle of $L^{\harmL}_{i^*_1}$ says ``non-member''. 

In case the adversary is dealing with case ($ii$), then the adversary simply picks $\trueL=L^{\trueL}_{i_1}$, $\harmL=L^{\harmL}_{i^*_1}$, and their enumerations $E^{\trueL}_{i_1}$ and $E^{\harmL}_{i^*_1}$ and the impossibility follows from the fact that $G$ never generates a string from $L^{\trueL}_{i_1}\setminus L^{\harmL}_{i^*_1}$.   
In case the adversary is dealing with case ($i$), then let $\hat{x}_1$ be the first string generated by the algorithm for which $\hat{x}_1\in L^{\trueL}_{i_1}$ and $\hat{x}_1\notin L^{\harmL}_{i^*_1}$, i.e., a safe string under the current setting. 
At this point, the adversary takes a conservative approach and assumes that the generator has the ability to safely generate from $L^{\trueL}_{i_1}$ when the harmful language is $L^{\harmL}_{i^*_1}$ (even if this is not necessarily true). 
For the remainder of this subphase, the adversary will take a few ``housekeeping'' actions to prepare for the next phase.

Suppose that string $E^{\trueL}_{i_1}(t_1)$ appears in position $t'_1$ in  $E^{\trueL}_{i^{\trueL}}$, i.e., the adversary has traversed up to position $t'_1$ in the original enumeration $E^{\trueL}_{i^{\trueL}}$ in order to generate the $t_1$-th string from  the enumeration $E^{\trueL}_{i_1}$ of this phase. 
Notice that $t'_1$ must be finite, i.e., $t'_1\neq \infty$. 
Define as $S^\trueL_1$ the strings from $E^{\trueL}_{i^{\trueL}}[1:t'_1]$ that are not in $E^{\trueL}_{i_1}[1:t_1]$, i.e., all the string from the original enumeration that were bypassed to pretend an enumeration from $L^{\trueL}_{i_1}$. 

An analogous argument is made for the harmful language. 
Suppose that string $E^{\harmL}_{i_1}(t_1)$ appears in position $t''_1$ in  $E^{\harmL}_{i^{\harmL}}$. 
Notice that $t''_1$ must be finite. 
Define as $S^\harmL_1$ the strings from $E^{\harmL}_{i^{\harmL}}[1:t''_1]$ that are not in $E^{\harmL}_{i_1}[1:t_1]$, i.e., all the string from the original enumeration that were bypassed to pretend an enumeration from $L^{\harmL}_{i_1}$. 

If $S^\trueL_1\neq\emptyset$ or $S^\harmL_1\neq\emptyset$, then we proceed to Subphase 1.B; otherwise, if $S^\trueL_1=\emptyset$ and $S^\harmL_1=\emptyset$, we proceed straight to Subphase 1.C.
We denote with $S^\trueL_{t_1}$ (resp. $S^\harmL_{t_1}$) the strings that have been enumerated so far, i.e., $E^{\trueL}_{i_1}[1:t_1]$ (resp. $E^{\harmL}_{i_1}[1:t_1]$ ).

\textbf{Subphase 1.B (Add Skipped Elements):}
In this subphase, the adversary enumerates all the leftover strings from $S^\trueL_1$ and $S^\harmL_1$ so that we are guaranteed that $G$ has seen every single string in sequences $E^{\trueL}_{i^{\trueL}}[1:t'_1]$ and $E^{\harmL}_{i^{\harmL}}[1:t''_1]$. 
This action guarantees no ``left-over'' strings in  $E^{\trueL}_{i^{\trueL}}$ and $E^{\harmL}_{i^{\harmL}}$ due to the (unsuccesful) attempt of the adversary to trick $G$ by enumerating as if the choice is $\trueL=L^{\trueL}_{i_1}$, $\harmL=L^{\harmL}_{i^*_1}$.
Proceed to Subphase 1.C.

\textbf{Subphase 1.C (Nothing Skipped; Issue element outside $L^{\trueL}_{i_1}$):} Recall that $G$ has seen all elements in $E^{\trueL}_{i^{\trueL}}[1:t'_1]$. 
Now the adversary needs to enumerate a string from the new true language that contradicts the hypothesis $\trueL=L^{\trueL}_{i_1}$ of generator $G$.

We first argue that there exists a string $y_1$ in $E^{\trueL}_{i^{\trueL}}[t'_1:\infty]$ that does not belong to $L^{\trueL}_{i_1}$, such that, when the adversary enumerates it, the generator $G$ will recognize that $\trueL \ne L^{\trueL}_{i_1}$. 
From property $P^*$, we know that 
$ L^{\trueL}_{i_1}\subsetneq L^{\trueL}_{i^{\trueL}}$, therefore there exists a string $y_1$ in $L^{\trueL}_{i^{\trueL}}$ that does not belong to $L^{\trueL}_{i_1}$. 
Since $E^{\trueL}_{i^{\trueL}}$ contains all strings in $L^{\trueL}_{i_1}$ and since $y_1$ has not been enumerated so far, it must be in $E^{\trueL}_{i^{\trueL}}[t'_1:\infty]$. 
Let $t_{y_1}>t'_1$ be the position of  $E^{\trueL}_{i^{\trueL}}$ that is equal to $y_1$. 
As a final step, the adversary enumerates all the strings in $E^{\trueL}_{i^{\trueL}}[t'_1+1:t_{y_1}]$ and proceeds to phase 2.

\textbf{Phase $l$.} For the $l$-th phase, we will use the fact that $P^*$ holds but this time, the $T$ that we use is $E^{\trueL}_{i^{\trueL}}[1:t_{y_l}]$. 
That is, let $T^{\trueL}=\{E^{\trueL}_{i^{\trueL}}[1:t_{y_l}]\}$ and $T^{\harmL}=\{E^{\harmL}_{i^{\harmL}}[1:t''_l]\}$. 
Then from $P^*$, there exists a $i_l\geq 1$ and an associated $i^*_l\geq 1$, such that (1)  $T^{\trueL}\subseteq L^{\trueL}_{i_l}$ and $L^{\trueL}_{i_l}$ is a proper subset of $L^{\trueL}_{i^{\trueL}}$, and (2) $T^{\harmL}\subseteq L^{\harmL}_{i^*_l}$ and $L^{\harmL}_{i^*_l}$ is a proper subset of $L^{\harmL}_{i^{\harmL}}$, and (3) $L^{\trueL}_{i_l}\setminus  L^{\harmL}_{i^*_l}$ has infinite cardinality. 
Additionally, notice that $L^{\trueL}_{i_l}$ is different from all $L^{\trueL}_{i_1},\ldots,L^{\trueL}_{i_{l-1}}$ because it contains string $y_l$ which does not appear in any of these languages. 

The arguments after this point are analogous to the ones in phase 1. 

\textbf{Subphase $l$.A:}
Consider an enumeration of $L^{\trueL}_{i_l}$ constructed by traversing the fixed enumeration $E^{\trueL}_{i^{\trueL}}$ and using only the elements of $L^{\trueL}_{i_l}$ that appear in it and transfer them to $E^{\trueL}_{i_l}$ in the same order. 
Consider an enumeration of $L^{\harmL}_{i^*_l}$ constructed by traversing the fixed enumeration $E^{\harmL}_{i^{\harmL}}$ and using only the elements of $L^{\harmL}_{i^*_l}$ that appear in it and transfer them to $E^{\harmL}_{i^*_l}$ in the same order. 

Consider two cases: ($i$) either there is some finite $t_l\in\mathbb{N}$ such that the generator $G$ will generate an element from $L^{\trueL}_{i_l}\setminus L^{\harmL}_{i^*_l}$, or ($ii$) there is no such $t_l\in\mathbb{N}$. 
In case the adversary is dealing with case ($ii$), then the adversary simply picks $\trueL=L^{\trueL}_{i_l}$, $\harmL=L^{\harmL}_{i^*_l}$, and their enumerations $E^{\trueL}_{i_l}$ and $E^{\harmL}_{i^*_l}$ and the impossibility follows from the fact that $G$ never generates a string from $L^{\trueL}_{i_l}\setminus L^{\harmL}_{i^*_l}$.   
In case the adversary is dealing with case ($i$), then let $\hat{x}_l$ be the first string generated by the algorithm for which $\hat{x}_l\in L^{\trueL}_{i_l}$ and $\hat{x}_l\notin L^{\harmL}_{i^*_l}$, i.e., a safe string under the current setting. 
At this point, the adversary takes a conservative approach and assumes that the generator has the ability to safely generate from $L^{\trueL}_{i_l}$ when the harmful language is $L^{\harmL}_{i^*_l}$ (even if this is not necessarily true). 

Suppose that string $E^{\trueL}_{i_1}(t_l)$ appears in position $t'_l$ in  $E^{\trueL}_{i^{\trueL}}$. 
Notice that $t'_l$ must be finite, i.e., $t'_l\neq \infty$. 
Define as $S^\trueL_l$ the strings from $E^{\trueL}_{i^{\trueL}}[1:t'_l]$ that are not in $E^{\trueL}_{i_1}[1:t_l]$, i.e., all the string from the original enumeration that were bypassed to pretend an enumeration from $L^{\trueL}_{i_l}$. 

An analogous argument is made for the harmful language. 
Suppose that string $E^{\harmL}_{i_l}(t_l)$ appears in position $t''_l$ in  $E^{\harmL}_{i^{\harmL}}$. 
Notice that $t''_l$ must be finite. 
Define as $S^\harmL_l$ the strings from $E^{\harmL}_{i^{\harmL}}[1:t''_l]$ that are not in $E^{\harmL}_{i_1}[1:t_l]$, i.e., all the string from the original enumeration that were bypassed to pretend an enumeration from $L^{\harmL}_{i_l}$. 

If $S^\trueL_l\neq\emptyset$ or $S^\harmL_l\neq\emptyset$, then we proceed to Subphase $l$.B; otherwise, if $S^\trueL_l=\emptyset$ and $S^\harmL_l=\emptyset$, we proceed straight to Subphase 1.C.
We denote with $S^\trueL_{t_l}$ (resp. $S^\harmL_{t_l}$) the strings that have been enumerated so far, i.e., $E^{\trueL}_{i_l}[1:t_l]$ (resp. $E^{\harmL}_{i_l}[1:t_l]$ ).

\textbf{Subphase $l$.B (Add Skipped Elements):}
In this subphase, the adversary enumerates all the leftover strings from $S^\trueL_l$ and $S^\harmL_l$ so that we are guaranteed that $G$ has seen every single string in sequences $E^{\trueL}_{i^{\trueL}}[1:t'_l]$ and $E^{\harmL}_{i^{\harmL}}[1:t''_l]$. 
This action guarantees no ``left-over'' strings in  $E^{\trueL}_{i^{\trueL}}$ and $E^{\harmL}_{i^{\harmL}}$ due to the (unsuccesful) attempt of the adversary to trick $G$ by enumerating as if the choice is $\trueL=L^{\trueL}_{i_l}$, $\harmL=L^{\harmL}_{i^*_l}$.
Proceed to Subphase $l$.C.

\textbf{Subphase $l$.C (Nothing Skipped; Issue element outside $L^{\trueL}_{i_1}$):} Recall that $G$ has seen all elements in $E^{\trueL}_{i^{\trueL}}[1:t'_l]$. 
Now the adversary needs to enumerate a string from the new true language that contradicts the hypothesis $\trueL=L^{\trueL}_{i_l}$ of generator $G$.

We first argue that there exists a string $y_l$ in $E^{\trueL}_{i^{\trueL}}[t'_l:\infty]$ that does not belong to $L^{\trueL}_{i_l}$, such that, when the adversary enumerates it, the generator $G$ will recognize that $\trueL \ne L^{\trueL}_{i_l}$. 
From property $P^*$, we know that 
$ L^{\trueL}_{i_l}\subsetneq L^{\trueL}_{i^{\trueL}}$, therefore there exists a string $y_l$ in $L^{\trueL}_{i^{\trueL}}$ that does not belong to $L^{\trueL}_{i_l}$. 
Since $E^{\trueL}_{i^{\trueL}}$ contains all strings in $L^{\trueL}_{i_l}$ and since $y_l$ has not been enumerated so far, it must be in $E^{\trueL}_{i^{\trueL}}[t'_l:\infty]$. 
Let $t_{y_l}>t'_l$ be the position of  $E^{\trueL}_{i^{\trueL}}$ that is equal to $y_l$. 
As a final step, the adversary enumerates all the strings in $E^{\trueL}_{i^{\trueL}}[t'_l+1:t_{y_l}]$ and proceeds to phase 2. 

\textbf{Inductive Argument.} 
As explained, we continue the construction of the target enumeration inductively. 
If there exists some phase $(l+1)$ such that Case~($ii$) (in Subphase~A) is activated, then the impossibility follows. 
Now, assume that Case~($ii$) is not activated for any phase $(l+1)\in \mathbb{N}$. 
In that case, the adversary revealed an enumeration of $L^{\trueL}_{i^{\trueL}}$ and $L^{\harmL}_{i^{\harmL}}$ such that $G$ does not safely generate consistently from $L^{\trueL}_{i^{\trueL}}\setminus L^{\harmL}_{i^{\harmL}}$ which is empty.
To see this, recall that at the time of transitioning to a new phase, the generator generated a string that is not $\epsilon$, which is not a safe generation when $\trueL=L^{\trueL}_{i^{\trueL}}$ and $\harmL=L^{\harmL}_{i^{\harmL}}$. 
Since there are infinitely many such phases, there are infinitely many such timesteps so the generator fails. 
The reason that the induction can proceed for $t$ to infinity is that $L^{\trueL}_{i^{\trueL}}$ has infinite cardinality; thus, we can have an infinite number of phases where the algorithm cannot safely generate. 
Then, in this case, the impossibility follows by setting the target language $\trueL=L^{\trueL}_{i^{\trueL}}$ and $\harmL=L^{\harmL}_{i^{\harmL}}$ and using the enumeration we have inductively constructed over all phases.
\end{proof}

\xhdr{Remark 1}
If ($i$) collections $\colltrue$ and $\collharm$ are identifiable, and ($ii$) $A$ has access to a set difference oracle, then $A$ can safely generate in the limit.

\subsection{A Tractable Case: Language Collections with Guaranteed Infinite Set Differences}
\label{subsec:tractable}

Before going into the details of the proofs below, we establish a few general principles for an algorithm that attempts to safely generate.

First, we mention that, given an enumeration $(E_{\trueL}, E_{\harmL})$ of strings from $\trueL$ and $\harmL$ presented to $A$ by the adversary, the algorithm $A$ can efficiently maintain two lists of hypotheses $K$ and $H$ that are \emph{consistent} 
with the enumeration. This maintenance proceeds in a manner analogous to the approach of \citet{DBLP:conf/nips/KleinbergM24}-- which will be referred to as \texttt{KM} in the following.

Second, recall that, \textit{in the limit}, the only ``interesting'' languages from $\colltrue$ and $\collharm$ are those that will remain consistent with respect to the enumeration of the chosen $K,H$. Thus, the only languages that $A$ needs to track are those that form a hierarchy under set inclusion\footnote{Recall that all languages in our collection are infinite.} with respect to $K$ and $H$.

Third, we establish that, in order to safely generate from $\trueL\setminus\harmL$, we need to at least be able to generate from $\trueL$ alone. As \texttt{KM} showed, this is possible. Given a list of consistent languages $K$, in order to generate in the limit from the correct $K$, a conservative but effective approach is to choose the \textit{smallest} consistent $K$. 
In fact, the \texttt{KM} algorithm guarantees exactly that: it maintains a list of all consistent languages $L_i \subseteq \ldots \subseteq L_z \subseteq \ldots L_j$, and chooses the ``smallest safe'' language (\texttt{KM} calls it $(t,m)$-critical language), proving that in the limit their algorithm will always generate from either the correct $K$ or from one of its subset languages.

Given that \texttt{KM} establishes that a viable but conservative strategy for generation from $\trueL$ corresponds to selecting the smallest consistent candidate, we pose the following question: What is a conservative choice of candidates when the goal is to generate from  $\trueL \setminus \harmL$ instead? 
We show that much like \texttt{KM}, for generating from $\trueL$ the conservative choice is to go with the ``smallest'' candidate, but to avoid all consistent harmful candidates, the conservative choice is to go with the \emph{largest} consistent $\harmL$.

\xhdr{New Setting} 
First, consider a slightly different setting, where we provide the algorithm with collections $\colltrue,\collharm$ between which all pairs of possible $\trueL \in \colltrue, \harmL \in \collharm$ have an infinite difference: i.e., $K_i \setminus H_j$ has infinite cardinality $\forall i,j \text{ with } K_i \in \colltrue \text{ and } H_j \in \collharm$. This additional condition circumvents the decidability issue in the set difference problem. 
In the following, we present a variation of the definition of safe generation $\mathcal{SG}^{\infty}$ in which every pair of a true language from $\colltrue$ with a harmful language from $\collharm$ has \emph{by construction} an infinite set difference.

\begin{definition}[\textbf{Safe Generation - Infinite Differences $\mathcal{SG}^{\infty}$}]
Fix some $\trueL$ from the language collection  $\colltrue$ and some $\harmL$ from the language collection  $\collharm$. All possible pairs of languages from $\colltrue$ and $\collharm$ are guaranteed to have an enumerable (infinite) difference, such that $\forall \trueL \in \colltrue, \forall \harmL \in \collharm: |\trueL \setminus \harmL|=\infty$.
In each step $t=1,2,3,\ldots$, the algorithm $A$ observes a labeled set $S_t$ such that each string is labeled $1$ if it is a member of $\trueL$ and $0$ if it is a member of $\harmL$. 
At each timestep, $A$ outputs a string $a_t$. 
We say that algorithm A \textbf{safely generates from $\trueL$ given a harmful $\harmL$ with infinite set difference in the limit} if there is some time $t'\in \mathbb{N}$ such that for all steps $t> t'$ the algorithm's guess $a_t$  belongs to $(\trueL\setminus\harmL)\setminus S_t$.
\end{definition}

\begin{theorem}
\label{th:easy}
Let $\colltrue, \collharm$ be language collections, such that $\forall \trueL \in \colltrue, \forall \harmL \in \collharm: |\trueL \setminus \harmL|=\infty$. 
There is an algorithm $A$ that, for any labeled enumeration of any language $\trueL\in \colltrue$ and $\harmL\in \collharm$, $A$ safely generates from $\trueL$ given a harmful $\harmL$ with infinite set difference in the limit.
\end{theorem}

\begin{proof}   
The proposed $A$ essentially runs an instance of the \texttt{KM} generation, for ``choosing'' a good candidate for $\trueL$. 
For $\harmL$, we will always select the consistent language with the smallest index. 
As the arbitrarily chosen enumerations proceed (through an approach similar to \texttt{KM}), the constructed algorithm $A$ maintains an ordered list of languages from $\colltrue$ consistent with the enumeration $E_K$, and always chooses the ``smallest'' (in the \texttt{KM} sense) as its guess for the correct $K$ (denote its guess by  $K_c$). 
In a similar manner, $A$ maintains an list of languages from $\collharm$ consistent with the enumeration $E_H$, but now always choose the one with the smallest index as its guess for the correct $H$ (denote its guess by $H_c$).\footnote{Subscript $c$  stands for "chosen". $K_c$ and $H_c$ need not have the same index in their respective collections.} 
We know that after some $t$, through the \texttt{KM} algorithm, the selected language $K_c$ will either be a subset of or the correct $K$.
We also know that after some $t'$, $A$ will consider the correct language $H$, and from this point on, it will remain within our list of consistent $H$s.
In this set of consistent $H$s will, in the limit, the language with the smallest index will either be $H$ itself or a superset language.
Now, we argue that for any $t^* > \max(t,t')$, the algorithm will safely generate. After $t^*$, we know that (a) algorithm $A$ will always choose some $K_c \subseteq \trueL$, and (b) we always choose some $H_c \supseteq \harmL$. Importantly, from the setting, we also know that their difference $K_c \setminus H_c$ will be infinite, and thus $A$ can, in finite time, find a string to return from it, which will also belong in $K\setminus H$.\footnote{The difference needs to be infinite, not simply non-empty. This guarantees that $A$ can find a string in $(K_c\setminus H_c) \setminus S_t$. 
If
$(K_c\setminus H_c)$
is non-empty but finite, the adversary could first enumerate it, thereby leaving $A$ with no strings to generate.} Thus, the proposed algorithm $A$ safely generates in the limit.
\end{proof}

\subsection{Conservative Generation Fails in $\mathcal{SG}$}
Our findings show that the two collections \emph{do not have to be identifiable} if one guarantees that all set differences have infinite cardinality. 
The tractability proof of $\mathcal{SG}^\infty$ shows that the conservative generation strategy of selecting the smallest candidate $K$ and the largest candidate $H$ is sound. 

In the general $\mathcal{SG}$ case, though, this conservative generation strategy can not be used. Consider again an algorithm that maintains two lists of languages $K_1 \subseteq \ldots \subseteq K_i \subseteq \ldots \subseteq K_n$ and $H_1 \subseteq \ldots \subseteq H_j \subseteq \ldots \subseteq H_m$ which are both consistent with the respective enumerations $E_K$ and $E_H$. For simplicity, assume that  $A$ is at a timestep where both the correct $K_i$ and $H_j$ are under consideration and are members of the two lists.
The conservative choice for the algorithm is still to select $K_c \leftarrow K_1$ and $H_c \leftarrow H_m$ (i.e., the smallest $K$ and the largest $H$). 
Equivalently, this strategy maximizes the portion removed from true $K$: it selects the smallest subset of $K$ and, in addition, removes the largest portion through (a potentially superset of) $H$. This gives the smallest remaining difference $K \setminus H$. 
Without Theorem's~\ref{th:easy} assumptions about the cardinality of set difference, though, the difference $K_c \setminus H_c$ could be empty (in which case $A$ is tempted to output $\perp$), even if the difference $K \setminus H$ is not empty, i.e., safe generation is possible.

\section{Conclusion}
Here, we discuss informal practical takeaways from our results.
If the class of harmful languages $\collharm$ is overly expansive, the resulting intersection with $\trueL$ may become large (even encompassing the entire $\trueL$), which is the primary source of intractability. In this sense, an overly generous definition of harmfulness can itself render the task intractable. 
 
The abstraction in this work is deliberately broad: it makes no assumptions about the internal learning strategy of $A$ or the structure of harmful languages in $\collharm$. 
For example, $\harmL$ could be as simple as \emph{``all strings containing the word \texttt{bomb}''}, which should be easy to avoid in generation. 
The generality of our abstraction comes at a cost: it may shift
attention away from more constrained settings in which safe generation is tractable.
 
One positive interpretation of our results is that \textit{narrow} definitions of safety are provably achievable in~practice. 
For example, safe generation is tractable if the harmful part of $\trueL$ (i.e., $\trueL\cap\harmL$) is finite, even if $\harmL$ itself is infinite.\footnote{We note that finiteness does not necessarily imply the existence of efficient algorithms.} 

Ultimately, our hope is that the community will develop the appropriate theory for real-world-driven challenges from deployed models (e.g., safety) to lay the foundations of language generation.

\section*{Acknowledgements}

We are thankful to Grigoris Velegkas for pointing out an error in our proof in a previous version of our manuscript.
Antonis Anastasopoulos is generously funded by the NSF under CAREER award 439202.

\bibliographystyle{iclr2026_conference}
\bibliography{biblio}

\clearpage
\appendix

\section{Description of the \texttt{KM} Algorithm}
\label{app:km}

\paragraph{Intuition and High-level Goal.}
The purpose of the algorithm of~\citet{DBLP:conf/nips/KleinbergM24} in Section 5.1 is to achieve \emph{generation in the limit} without ever identifying the true language. The algorithm observes only positive examples
\[
S_t = \{w_1,\ldots,w_t\}
\]
from an unknown target language \(K = L_z\), drawn from a countable list of candidate languages
\[
\mathcal{C} = \{L_1, L_2, L_3, \ldots\}.
\]
The algorithm has access to a membership oracle to query whether a string belongs to a candidate language \(L_i\), but it cannot test membership in the unknown target language $K$ directly. Since language identification is provably impossible in this setting, the algorithm instead adopts a \emph{conservative} strategy: it never commits to a single hypothesis for \(K\), but instead generates strings that are guaranteed to lie in \emph{all} plausible hypotheses consistent with the data seen so far. 

\paragraph{The Role of \((t,m)\)-critical Languages.}
Exact set containment between languages cannot be tested in finite time due to the fact that the languages are infinite. 
To overcome the testing membership problem for an infinite number of strings, \texttt{KM} tests containment on a \emph{restricted part of the vocabulary}; this approach can be viewed as an approximation of the (infinite) containment computation. 
On a more technical note, the universe of strings is enumerated as
\[
U = \{u_1, u_2, u_3, \ldots\},
\]
and for (1) each language \(L_i\) and (2) cutoff \(m\), the algorithm considers the finite prefix (i.e., the restricted vocabulary)
\[
L_i[m] = L_i \cap \{u_1,\ldots,u_m\}.
\]
At step \(t\), the algorithm restricts attention to the first \(t\) candidate languages \(\mathcal{C}_t = \{L_1,\ldots,L_t\}\) and defines a language \(L_n \in \mathcal{C}_t\) to be \emph{\((t,m)\)-critical} if it is consistent with the observed sample \(S_t\), and if its finite prefix \(L_n[m]\) is contained in the prefix of every consistent language that appears before $L_n$ in $\mathcal{C}_t$. 
Intuitively, a \((t,m)\)-critical language is the "smallest" consistent language given the current finite evidence on this bounded vocabulary (controlled by $m$) and bounded subcollection of languages (controlled by $t$). 

\paragraph{Generation Procedure and Convergence.}
In each step \(t\), the algorithm begins with a cutoff \(m\) large enough to include all strings observed so far, and then incrementally increases \(m\). For each value of \(m\), it identifies the highest-indexed \((t,m)\)-critical language \(L_{n_t(m)}\) (here, the $n_t(m)$ variable denotes the highest-indexed language among those that are $(t,m)$-critical.). The algorithm then searches for the first string \(u_i \le m\) such that
\[
u_i \in L_{n_t(m)} \setminus S_t.
\]
If such a string exists, it is output and the generation of this timestep concludes; otherwise, \(m\) is increased and the process repeats. The new $m$ can lead to a different language becoming $(t,m)$-critical. 
This search is guaranteed to terminate, i.e., the identity of the highest-indexed \((t,m)\)-critical language can change only finitely many times as \(m\) grows, and once it stabilizes, the ``infiniteness'' of the language's cardinality ensures the existence of a new string. Moreover, after some finite time \(t^*\), the true language \(L_z\) becomes \((t,m)\)-critical for all \(m\), implying that every subsequently selected language is either $L_z$ or a subset of \(L_z\). 
From that point onward, every generated string lies in \(K \setminus S_t\). The algorithm thus succeeds by progressively shrinking the space of safe outputs until it is guaranteed to lie entirely within the true language, without ever identifying it.

\end{document}